%% file: main.tex
\documentclass[]{academic_template}

\usepackage[toc,page,header]{appendix}

\usepackage{amssymb}
\usepackage{amsmath}

\usepackage{makecell}
\usepackage{algorithm}      
\usepackage{algpseudocode}  

\usepackage{graphicx}

\usepackage{listings}
\usepackage{url}
\usepackage{booktabs}
\usepackage{wrapfig}
\newcommand{\tablestyle}[2]{\setlength{\tabcolsep}{#1}\renewcommand{\arraystretch}{#2}\centering\footnotesize}

\usepackage{colortbl}
\usepackage{xcolor}
\usepackage[dvipsnames]{xcolor}

\usepackage{caption}
\usepackage{subcaption}

\tcbset{
  aibox/.style={
    top=10pt,
    colback=white,
    colframe=black,
    colbacktitle=black,
    enhanced,
    center,
    breakable,
    attach boxed title to top left={yshift=-0.1in,xshift=0.15in},
    boxed title style={boxrule=0pt,colframe=white,},
  }
}
\newtcolorbox{AIbox}[2][]{aibox, title=#2,#1}

\setlogoheight{14mm}   
\setlogospacing{5mm}
\setsjtublue 

\settitlerulethickness{3pt}  

\abstractboxon 
\setabstractframecolor{gray} 
\setabstractbgcolor{gray!10} 
\setlogotolineshift{5mm} 

\settoprulethickness{2.5pt}  
\setbottomrulethickness{1.5pt}

\title{Trust Your Critic: Robust Reward Modeling and Reinforcement Learning for Faithful  Image Editing and Generation}

\author[1*]{Xiangyu Zhao}
\author[2*]{Peiyuan Zhang}
\author[3*]{Junming Lin}
\author[6*]{Tianhao Liang}
\author[4]{Yuchen Duan}
\author[4]{Changyao Tian}
\author[5,6]{Shengyuan Ding}
\author[6]{Yuhang Zang}
\author[1]{Junchi Yan}
\author[1, \dagger]{Xue Yang}

\affiliation[1]{Shanghai Jiao Tong University}
\affiliation[2]{Wuhan University}
\affiliation[3]{BUPT}
\affiliation[4]{CUHK}
\affiliation[5]{Fudan University}
\affiliation[6]{Shanghai AI Laboratory}

\contribution[*]{Equal contribution}
\contribution[\dagger]{Corresponding Author}

\abstract{
Reinforcement learning (RL) has emerged as a promising paradigm for enhancing image editing and text-to-image (T2I) generation. However, current reward models, which act as critics during RL, often suffer from hallucinations and assign noisy scores, inherently misguiding the optimization process. 
In this paper, we present \textbf{FIRM} (\textbf{F}aithful \textbf{I}mage \textbf{R}eward \textbf{M}odeling), a comprehensive framework that develops robust reward models to provide accurate and reliable guidance for faithful image generation and editing. 
First, we design tailored data curation pipelines to construct high-quality scoring datasets. Specifically, we evaluate editing using both execution and consistency, while generation is primarily assessed via instruction following. Using these pipelines, we collect the FIRM-Edit-370K and FIRM-Gen-293K datasets, and train specialized reward models (FIRM-Edit-8B and FIRM-Gen-8B) that accurately reflect these criteria. 
Second, we introduce FIRM-Bench, a comprehensive benchmark specifically designed for editing and generation critics. Evaluations demonstrate that our models achieve superior alignment with human judgment compared to existing methods. 
Furthermore, to seamlessly integrate these critics into the RL pipeline, we formulate a novel "Base-and-Bonus" reward strategy that balances competing objectives: Consistency-Modulated Execution (CME) for editing and Quality-Modulated Alignment (QMA) for generation. 
Empowered by this framework, our resulting models FIRM-Qwen-Edit and FIRM-SD3.5 achieve substantial performance breakthroughs. Comprehensive experiments demonstrate that FIRM mitigates hallucinations, establishing a new standard for fidelity and instruction adherence over existing general models. All of our datasets, models, and code will be publicly available.
}

\date{\today}

\checkdata[Project Page]{\url{https://firm-reward.github.io/}}
\checkdata[Code]{\url{https://github.com/VisionXLab/FIRM-Reward}}
\checkdata[Hugging Face]{\url{https://huggingface.co/collections/VisionXLab/firm-reward}}
\begin{document}
\maketitle


\input{sections/1_introduction}

\input{sections/2_related_work}
\input{sections/3_method}
\input{sections/4_experiments}
\input{sections/5_conclusion}


\bibliographystyle{plainnat}
\bibliography{references}

\clearpage
\beginappendix
\input{sections/appendix}

\end{document}

%% file: sections/1_introduction.tex
\section{Introduction}
The rapid advancement of diffusion models and autoregressive models has revolutionized both text-to-image (T2I) generation~\cite{crowson2022vqgan, balaji2022ediff,zhang2023adding,saharia2022photorealistic} and image editing~\cite{liu2025step1x, wu2025qwenimagetechnicalreport,zhang2025icedit}. Recently, Reinforcement Learning (RL) has emerged as a prevailing paradigm, relying heavily on reward models (or "critics") to provide optimization signals. In this RL-driven alignment process, the quality of the generated or edited images is fundamentally bound by the accuracy and reliability of the critics. 
Despite the widespread adoption of RL, current pipelines face a critical bottleneck: the unreliability of the critics. While recent Multimodal Large Language Models (MLLMs)~\cite{gpt5, Google_Gemini3Pro_2025, bai2025qwen3, team2026kimi, wang2025internvl3} have demonstrated impressive general capabilities, they frequently struggle when employed as zero-shot reward models for fine-grained image editing and generation tasks. These models inherently suffer from severe hallucinations, object neglect, and a lack of precise spatial reasoning, leading to unreasonable and noisy reward scores. 

\begin{figure}[t]
    \centering
    \includegraphics[width=\linewidth]{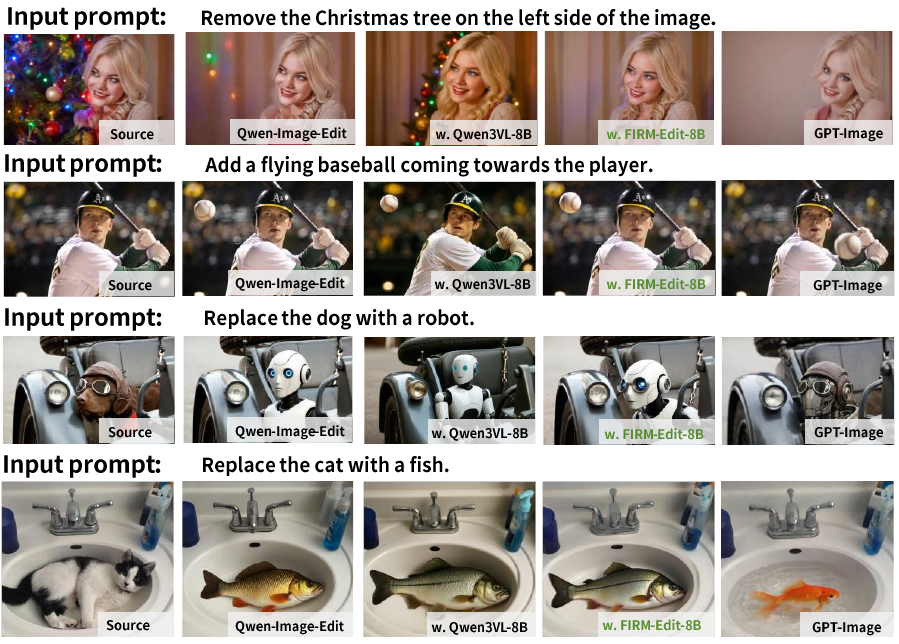}
    \caption{Comparison of image editing results across different methods. “w. FIRM-Edit-8B” indicates that FIRM-Edit-8B is adopted as the reward model during RL process.}
    \label{fig:edit_case}
\end{figure}

To address these critical limitations, we present \textbf{FIRM} (\textbf{F}aithful \textbf{I}mage \textbf{R}eward \textbf{M}odeling), a comprehensive framework designed to train robust, task-specific reward models that serve as trustworthy critics for both image editing and generation. FIRM introduces tailored data construction pipelines to synthesize high-quality reward data. For image editing, we observe a crucial counter-intuitive phenomenon: while MLLMs struggle to directly judge whether an edited image perfectly follows instructions while preserving consistency, they excel at identifying the differences between two images. Therefore, our editing pipeline adopts a "difference-first" approach. We leverage an MLLM to caption the exact visual differences between the source and edited images, and subsequently feed this textual difference to an MLLM to reliably deduce the final scores for both execution and consistency. For T2I generation, we propose a checklist-based prompting strategy: an LLM first extracts key checking points from the user prompt, which are then appended to the MLLM's scoring prompt. This explicitly guides the MLLM to verify fine-grained details, significantly reducing hallucinations. By utilizing the pipelines and existing open-source datasets, we construct two high-quality reward datasets: FIRM-Edit-370K and FIRM-Gen-293K. Subsequently, we employ these datasets to train our reward models, FIRM-Edit-8B and FIRM-Gen-8B, which are initialized from the Qwen3-VL-8B-Instruct model.

\begin{figure}[t]
    \centering
    \includegraphics[width=.9\linewidth]{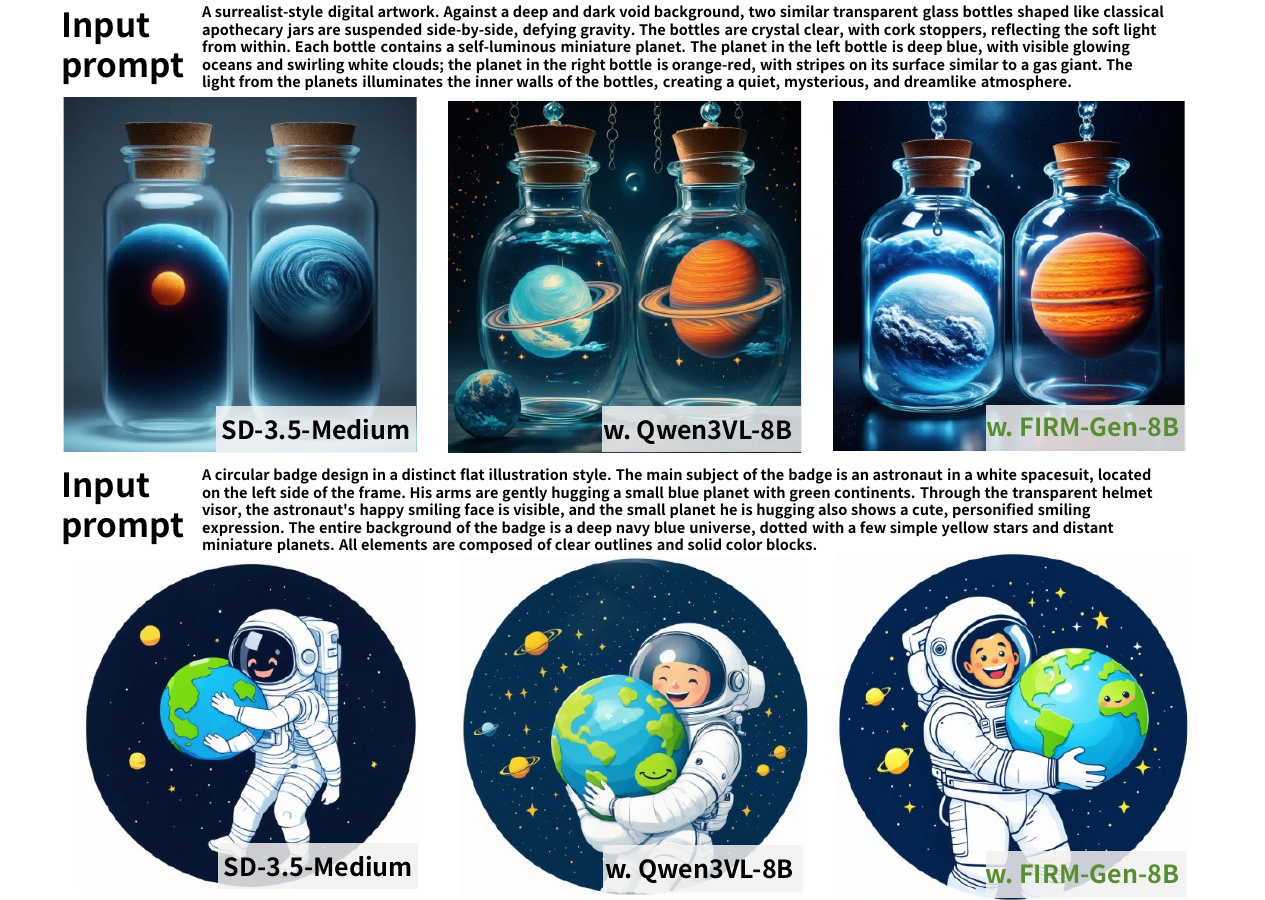}
    \caption{Comparison of T2I generation results across different methods. “w. FIRM-Gen-8B” indicates that FIRM-Gen-8B is adopted as the reward model during RL process.}
    \label{fig:gen_case}
\end{figure}

To rigorously validate our critics, we also construct a human-annotated benchmark FIRM-Bench for both generation and editing tasks. By sourcing prompts from diverse existing benchmarks and images from various models while strictly controlling the ground-truth score distribution, we demonstrate that our trained reward models achieve remarkable alignment with human preference, vastly outperforming existing open-source MLLMs. 

Building upon these trustworthy critics, we further perform RL to optimize generative models. A well-known challenge in RL is that naively maximizing multiple, often competing rewards frequently triggers optimization collapse or severe reward hacking. To mitigate this, we extensively explore "Base-and-Bonus" reward weighting strategies and identify a synergistic reward formulation termed Consistency-Modulated Execution (CME). This strategy successfully balances the trade-off and prevents reward hacking in image editing. Similarly, for T2I generation, we propose Quality-Modulated Alignment (QMA) to balance instruction following and image quality. Guided by our robust critics and formulations, we successfully train the FIRM-Qwen-Edit and FIRM-SD3.5 models. Extensive experiments demonstrate that our models yield substantial performance gains across both generation and editing paradigms, as illustrated in \cref{fig:edit_case} and \cref{fig:gen_case}.

In summary, the main contributions of this work are four-fold:

1. We propose the FIRM Framework, which consists of two specialized data construction pipelines (FIRM-Gen and FIRM-Edit) to train robust reward models. We design a novel "difference-first" (MLLM-to-LLM) pipeline for image editing and a checklist-guided approach for T2I generation. These pipelines yield the high-quality reward datasets FIRM-Edit-370K and FIRM-Gen-293K, along with their corresponding reward models, FIRM-Edit-8B and FIRM-Gen-8B.

2. We construct a comprehensive, fully human-annotated benchmark FIRM-Bench for both image editing and generation. Evaluations confirm that our reward models achieve superior alignment with human compared to existing models.

3. We propose a novel "Base-and-Bonus" reward fusion strategy in RL process, encompassing Consistency-Modulated Execution (CME) for editing and Quality-Modulated Alignment (QMA) for generation. This formulation effectively mitigates reward hacking and ensures balanced performance across various scenes.

4. Through extensive RL experiments yielding FIRM-Qwen-Edit and FIRM-SD3.5 models, we demonstrate that FIRM reward models and reward strategies lead to substantial and consistent performance enhancements in both faithful image editing and precise image generation, validating the efficacy of our overall approach.

%% file: sections/2_related_work.tex
\section{Related Works}

\subsection{Recent development of Image Editing and Generation}
Text-to-image (T2I) generation has witnessed a fundamental paradigm shift from early adversarial methods like GANs\cite{goodfellow2014generative}, and variational approaches like VAEs\cite{kingma2013auto}, to diffusion-based architectures\cite{ho2020denoising, nichol2021glide, rombach2022high}. Consequently, flow-based models\cite{esser2024scaling, flux2024} have emerged as an efficient alternative, offering accelerated sampling without compromising synthesis quality. Autoregressive models\cite{yu2022scaling, tian2024visual, xie2024showo} have also gained attention by treating image synthesis as a sequence modeling task. 

Building upon the capabilities of generation, image editing has rapidly evolved. Initial diffusion-based models~\cite{hertz2022prompt, wallace2023edict} heavily relied on dual-prompt formulations. 
Pioneered by InstructPix2Pix\cite{brooks2023instructpix2pix}, this trajectory has been substantially refined by recent works\cite{zhang2023magicbrush, sheynin2024emu} through the curation of large-scale, high-quality datasets. More recently, flow-matching models\cite{labs2025flux1kontextflowmatching} are enhancing training and sampling efficiency, while sequential autoregressive approaches\cite{yu2022scaling, tian2024visual} are fundamentally improving compositional reasoning. At the forefront of this evolution are hybrid multimodal architectures, such as BAGEL\cite{deng2025emerging} and Qwen-Image\cite{wu2025qwenimagetechnicalreport}. 

\subsection{Reinforcement Learning for Image Editing and Generation}

Traditional diffusion models~\cite{rombach2022high, podell2023sdxl} are primarily optimized via maximum likelihood estimation (MLE) to match the underlying data distribution. Recently, Reinforcement Learning (RL) has emerged as a pivotal paradigm. Early milestone approaches, such as DDPO~\cite{black2023training} and DPOK~\cite{fan2023dpok}, formulate the iterative denoising process of diffusion models as a multi-step Markov Decision Process (MDP), allowing for direct policy optimization via Proximal Policy Optimization (PPO).

Recently, the image generation field has integrated Chain-of-Thought (CoT) reasoning with RL. T2I-R1~\cite{jiang2025t2i} proposes a bi-level CoT reasoning framework optimized via Group Relative Policy Optimization (GRPO). 
EDIT-R1~\cite{li2025uniworld} explicitly tackles the absence of a universal editing reward by utilizing an MLLM as a unified, training-free reward model, leveraging its output logits to provide fine-grained feedback.
However, in image generation and editing, general-purpose MLLMs frequently fail to provide reliable rewards. EditScore~\cite{luo2025editscore} introduces a rigorous benchmarking suite and a family of high-fidelity reward models that significantly outperform open-source MLLMs.
EditReward~\cite{wu2025editreward} used a human-annotated dataset to fine-tune MLLM for data filtering. The methods above lack a comprehensive study of scalable data curation for reward models in both image generation and editing.

%% file: sections/3_method.tex
\section{Method}

\subsection{FIRM-Edit Pipeline}
In practice, we observed a notable phenomenon: models typically perform better as problem-solvers than as evaluators. Specifically, when tasked with judging an edited image,
models frequently fail to capture fine-grained details that they could otherwise identify in a "solving" (e.g., descriptive) context. This discrepancy leads to an overall mismatch in final evaluation scores. 

\begin{figure}[t!]
    \centering
    \includegraphics[width=\linewidth]{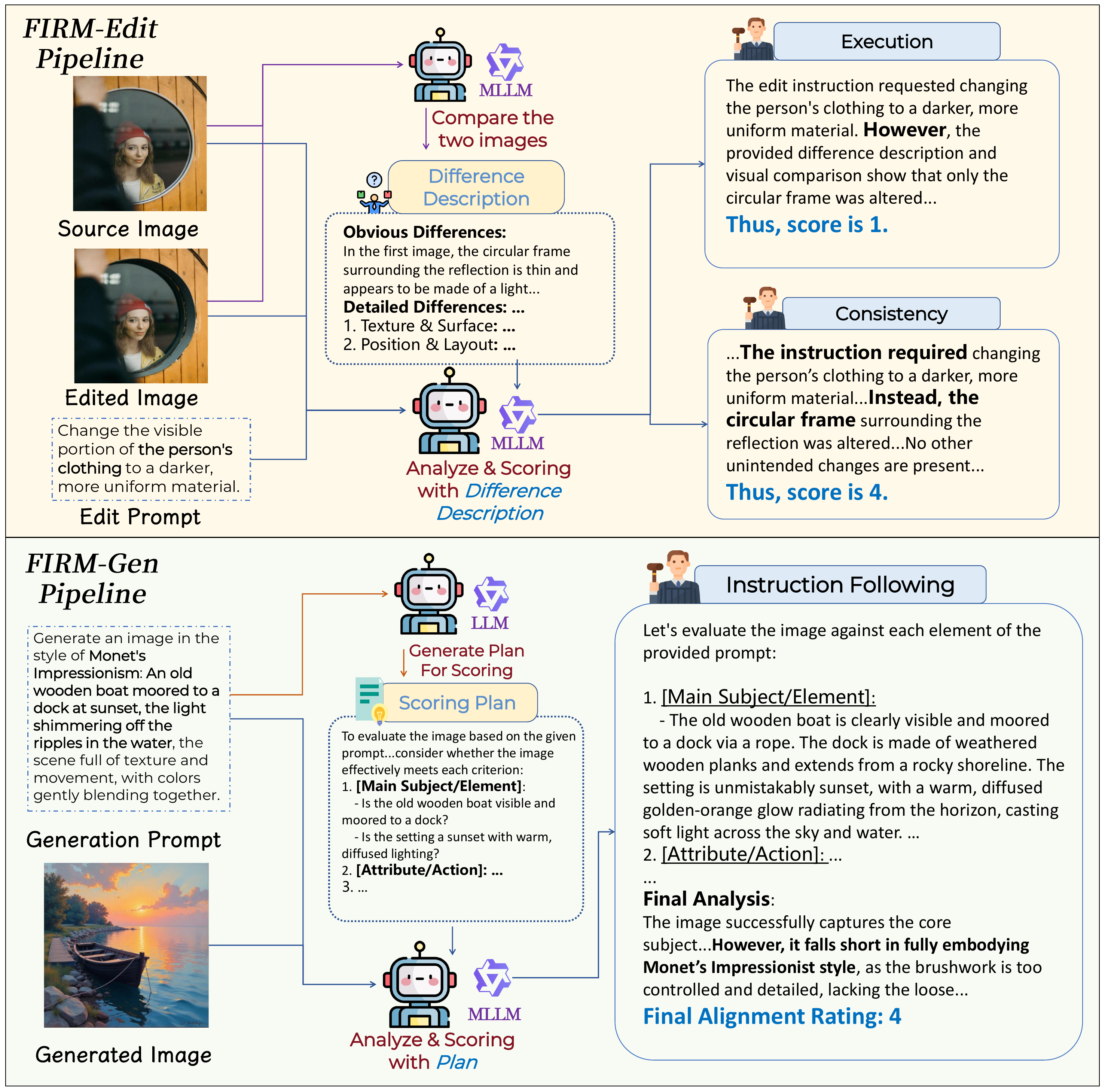}
    \caption{
    Overview of the FIRM data curation pipelines. (Top) The FIRM-Edit pipeline follows a novel "difference-first" design. (Bottom) The FIRM-Gen pipeline adopts a "plan-then-score" paradigm to significantly enhance scoring accuracy.
    }
    \label{fig:pipeline_compare}
\end{figure}

To address this, we introduce the FIRM-Edit pipeline, a "difference-first" methodology designed to generate a large-scale, high-quality VQA dataset for edit evaluation, as shown in \cref{fig:pipeline_compare}. First, given the original and edited image pairs, we prompt a state-of-the-art (SOTA) MLLM to perform a dual-level difference analysis. Specifically, the model is instructed to identify both obvious and detailed modifications, synthesizing them into a unified difference report. The generated comparative descriptions can effectively capture minor alterations made during the editing process, yielding a comprehensive observation. Subsequently, this description, alongside the image pair and the original edit instruction, is fed into a robust MLLM acting as the evaluator. By explicitly conditioning on this change description, the MLLM evaluator yields a final assessment that aligns significantly better with human expert judgments.

Regarding the reward formulation, inspired by RISEBench~\cite{zhao2025envisioning}, we assess the editing performance across two primary dimensions: execution and consistency, with both scores ranging from 1 to 5. Execution quantifies the accuracy of the model in executing the user's prompt, whereas consistency measures the preservation of unmodified objects and regions. An optimal edit must successfully satisfy all editing constraints while maintaining the integrity of the original context. By decoupling the evaluation process into these two distinct dimensions, our approach provides fine-grained reward signals, thereby facilitating superior model performance.
The source data is curated from publicly available image editing datasets, including OpenGPT-4o-Image~\cite{chen2025opengpt}, GPT-Image-Edit~\cite{wang2025gpt}, ShareGPT-4o-Image~\cite{chen2025sharegpt}, and ImgEdit~\cite{ye2025imgedit}. Since low-scoring examples (e.g., scores of 1 or 2) are extremely scarce in training datasets, we intentionally rewrite a subset of the instructions to synthetically generate poor-quality matches, ensuring a balanced distribution of reward scores.

\subsection{FIRM-Gen Pipeline}

Similar to the challenges observed in image editing, directly prompting an MLLM to assign a holistic score to a generated image often results in coarse-grained evaluations. When evaluating T2I generation, especially complex instructions with a large number of requirements, current open-source models struggle to simultaneously weigh multiple complex constraints (e.g., entity counts, spatial relationships, and stylistic attributes), leading to misaligned or unexplainable reward signals. Just as a professional photography judge first comprehends the evaluation objective, identifies key visual dimensions, and analyzes them individually before summarizing an overall score, we argue that a robust T2I reward model requires an explicit structural breakdown of the evaluation criteria.
Therefore, we introduce our FIRM-Gen pipeline (as shown in \cref{fig:pipeline_compare}), a "plan-then-score" methodology designed to elicit precise and interpretable judgments. 

FIRM-Gen pipeline decomposes the assessment process into two distinct stages.
In the first stage (Explicit Criteria Planning), we utilize a powerful LLM (Qwen3-32B\cite{bai2025qwen3}) acting as the "planner." Given the original text-to-image prompt, the planner is tasked with formulating a detailed, customized scoring checklist. This plan dynamically breaks down the prompt into fine-grained evaluation dimensions, such as main subject/element accuracy, style/composition alignment, and potential negative constraints.
In the second stage (Structured Analytical Scoring), a strong MLLM (Qwen3-VL-235B-A22B\cite{bai2025qwen3}) takes on the role of the "evaluator." It receives the generated image, the original generation prompt, and the customized scoring plan derived from the first stage. By explicitly conditioning the visual analysis on this structured plan, the MLLM is forced to conduct a step-by-step inspection of each predefined dimension before aggregating them into a final score. This decouple-and-conquer strategy effectively mitigates the "attention dilution" problem in MLLMs, enabling more robust, accurate, and explainable reward modeling that closely mimics human cognitive processes.

To ensure the trained reward model generalizes well across diverse scenarios, we construct a highly comprehensive source dataset. The generation prompts are sampled from mainstream, high-quality datasets, including OpenGPT-4o-Image\cite{chen2025opengpt}, ShareGPT-4o-Image\cite{chen2025sharegpt}, and BLIP3o-60k\cite{chen2025blip3}, covering a wide spectrum of user intents and complexities. Furthermore, to ensure the training data covers a wide range of image qualities and generation styles, we use a diverse set of models to generate images from these prompts. The pool incorporates models with varying architectures and capacities, specifically Ovis-image\cite{wang2025ovis_image}, Z-image-turbo\cite{team2025zimage}, Flux.1-dev\cite{flux2024}, SDXL\cite{podell2023sdxl}, and SD1.5\cite{rombach2022high}. This deliberate diversity prevents the reward model from overfitting to the artifacts of a single generator, thereby establishing a universal and reliable reward signal for T2I generation.

\subsection{Construction of FIRM-Bench}

To rigorously validate our reward models, we introduce FIRM-Bench, a comprehensive benchmark encompassing 807 meticulously curated samples. The benchmark is partitioned into two subsets: FIRM-Bench-Edit (301 for execution and 256 for consistency) and FIRM-Bench-Gen (250 for instruction following).

\noindent\textbf{Data Collection and Annotation.} To circumvent data contamination, we strictly isolate our benchmark from existing training datasets. We sample prompts from standard benchmarks and collect the corresponding result images from the open-source outputs of current popular models. Subsequently, human experts are employed to score each prompt-image pair according to the task.

\noindent\textbf{Dataset Balancing and Evaluation Protocol.} To ensure a robust and unbiased evaluation, we carefully control the sampling process to maintain a uniform distribution of human-annotated ground-truth scores (ranging from 1 to 5) across all metrics. Furthermore, we stratify FIRM-Bench-Gen into "easy" and "hard" subsets based on instruction complexity, enabling a granular analysis of model capabilities. Ultimately, the performance of a given reward model is quantified using the Mean Absolute Error (MAE) between the model's predicted scores and the human-annotated ground truth.

\subsection{Rewards Design in RL}
\textbf{Online RL Algorithm.}
We optimize our model using DiffusionNFT~\cite{zheng2025diffusionnft}, an online RL paradigm defined on the \emph{forward} diffusion process via flow matching.
Let the forward noising process be:
\begin{equation}
  x_t = \alpha_t x_0 + \sigma_t \epsilon,\quad \epsilon \sim \mathcal{N}(0, I)
\end{equation}
and let the velocity parameterization be trained by flow matching:
\begin{equation}
  \mathbb{E}_{c,\,x_0\sim \pi_{\text{old}}(\cdot|c),\,t,\,\epsilon}\big[w(t)\lVert v_\theta(x_t, c, t) - v \rVert_2^2\big],
\end{equation}
where the target velocity is determined by the noise schedule derivatives:
\begin{equation}
  v = \dot{\alpha}_t x_0 + \dot{\sigma}_t \epsilon
\end{equation}
DiffusionNFT defines a reinforcement-guided target policy in the velocity field as:
\begin{equation}
  v^\ast(x_t, c, t) = v_{\text{old}}(x_t, c, t) + \frac{1}{\beta}\Delta(x_t, c, t),
\end{equation}
where $\Delta(x_t,c,t)$ is the \emph{reinforcement guidance} direction derived from the contrast between
the positive and negative splits of $\pi_{\text{old}}$. 

Given an \emph{optimality probability} $r\in[0,1]$ for each sampled clean image $x_0$, DiffusionNFT optimizes the negative-aware objective:
\begin{equation}
  \mathcal{L}(\theta)
  = \mathbb{E}_{c,\,x_0\sim \pi_{\text{old}}(\cdot|c),\,t,\,\epsilon}
  \Big[r\lVert v_\theta^{+}(x_t,c,t)-v\rVert_2^2
  +(1-r)\lVert v_\theta^{-}(x_t,c,t)-v\rVert_2^2\Big]
\end{equation}
with implicit positive/negative policies
\begin{equation}
  v_\theta^{+}(x_t,c,t)=(1-\beta)v_{\text{old}}(x_t,c,t)+\beta v_\theta(x_t,c,t)
\end{equation}
\begin{equation}
  v_\theta^{-}(x_t,c,t)=(1+\beta)v_{\text{old}}(x_t,c,t)-\beta v_\theta(x_t,c,t)
\end{equation}
Under unlimited data and model capacity, the optimal solution of the above objective satisfies
\begin{equation}
  v_\theta^\ast(x_t,c,t)=v_{\text{old}}(x_t,c,t)+\frac{2}{\beta}\Delta(x_t,c,t)
\end{equation}

\noindent \textbf{Reward Formulation for Image Editing.} Within the RL framework, we design a reward method composed of two complementary signals: \textit{Execution} and \textit{Consistency}, each scored on a 1--5 scale and normalized to $[0,1]$ during training. 
However, a key challenge lies in proper credit assignment between these two signals. Our initial design employed a simple linear combination:
\[
R = w_1 \cdot \text{Consistency} + w_2 \cdot \text{Execution},
\]
where $w_1 = w_2 = 0.5$. This formulation led to severe reward hacking. In practice, the model discovered that maximizing Consistency is significantly easier than improving Execution. As a result, it converged to a degenerate strategy of outputting images nearly identical to the input, achieving high Consistency scores while failing to perform meaningful edits. This behavior severely hindered the improvement of editing capability under RL.

We first attempted to alleviate this issue by rebalancing the weights (e.g., increasing the Execution coefficient to 0.6). Although this adjustment slightly mitigated the problem, it did not fundamentally prevent the shortcut strategy.
To address this issue in a more principled manner, we adopted a more sensible reward function \textit{Consistency-Modulated Execution (CME)}:
\[
R_{CME} = \text{Execution} \cdot \left(w_1 + w_2 \cdot \text{Consistency}\right),
\]
where $w_1 = 0.6$ and $w_2 = 0.4$.
This formulation enforces Execution as a necessary condition for obtaining high reward: if Execution is low, the overall reward remains suppressed regardless of Consistency. Meanwhile, Consistency still acts as a shaping signal that refines structural fidelity once meaningful edits are performed. Empirically, this design leads to significantly better credit assignment, effectively prevents reward hacking, and substantially improves editing performance.

\noindent\textbf{Reward Formulation for Image Generation.} In our initial RL pipeline, we utilized Instruction Following as the exclusive reward signal. However, this naive setup exposed an intriguing instance of reward hacking. While the policy behaved as expected for comprehensive, highly detailed prompts, it discovered a trivial solution for short prompts comprising only bare object categories. Specifically, the model tended to synthesize a mere black shadow of the requested objects—perfectly satisfying the textual condition but entirely lacking intrinsic visual fidelity.
To address this, we posit that generation quality must act as a constraint. Drawing inspiration from our editing reward design, we propose \textit{Quality-Modulated Alignment (QMA)}:
\[
R_{QMA} = \text{InsFollowing} \cdot \left(w_1 + w_2 \cdot \text{Quality}\right),
\]
where $w_1 = 0.4$ and $w_2 = 0.6$. This reward shaping strategy places greater emphasis on image quality once the instruction-following score reaches a high level, effectively mitigating the aforementioned hacking behavior.

  

%% file: sections/4_experiments.tex
\section{Experiment Results}

\subsection{Experiment Settings}
For the Supervised Fine-Tuning (SFT) stage of the reward models, we initialize our FIRM-Edit-8B and FIRM-Gen-8B from Qwen3-VL-8B-Instruct model. They are trained on FIRM-Edit-370k and FIRM-Gen-293k datasets respectively. The SFT process is conducted on 8$\times$H200  GPUs using LLaMA-Factory~\cite{zheng2024llamafactory} codebase. 

To further validate the effectiveness of FIRM-Edit-8B and FIRM-Gen-8B, we integrate them as reward models to guide the reinforcement learning (RL) process of base models for image editing and generation tasks. 
During the RL stage, we employ the Edit-R1~\cite{li2025uniworld} framework for image editing and the Diffusion-NFT~\cite{zheng2025diffusionnft} framework for image generation. Each RL phase is scaled across 16$\times$H200 GPUs. Regarding the hyperparameters, we set the number of samples per rollout to $N=16$ and the overall batch size to $16$ (Editing) and $48$ (Generation). The training prompts for both the editing and generation RL processes are sourced from the ShareGPT-4o-Image~\cite{chen2025sharegpt} dataset. 
In RL phase, we train the editing model for 150 steps and the generation model for 600 steps.
We benchmark the post-RL performance against current state-of-the-art (SOTA) methods. Additionally, to account for both efficiency and capability, we conduct an ablation study by replacing our reward models with Qwen3-VL-8B and Qwen3-VL-32B.
\begin{figure}[t]
    \centering
    \includegraphics[width=\linewidth]{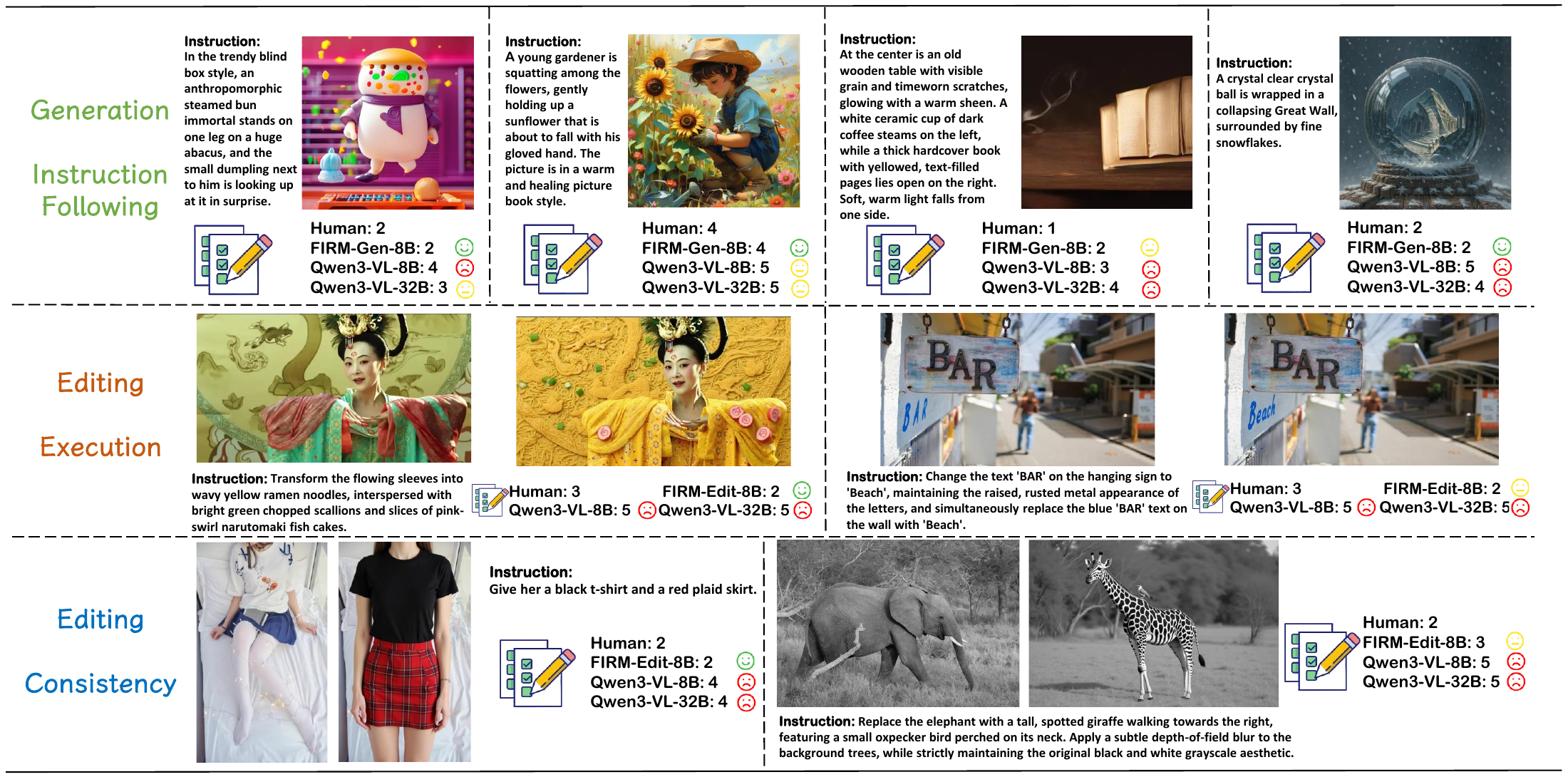}
    \caption{Illustrative examples from FIRM-Bench, showing that our reward models
    are better aligned with human judgments than Qwen3-VL-8B and Qwen3-VL-32B.}
    \label{fig:bench_cases}
\end{figure}

\input{sections/tabs/tab_main_bench}

\subsection{Evaluation Results on FIRM-Bench}
We conduct experiments evaluating several current state-of-the-art (SOTA) models on FIRM-Bench, including proprietary models (the GPT series and Gemini) and open-source models (the Qwen3-VL and InternVL-3.5 series). The results are presented in \cref{tab:bench_generation} and \cref{tab:bench_edit}. We show more cases in \cref{fig:bench_cases}.

On FIRM-Bench-Edit, the performance gap between proprietary and open-source models widens further. Gemini-3-Pro yields the second-lowest MAE of 0.54 on Execution and the lowest MAE of 0.57 on Consistency, resulting in an overall MAE of 0.55. In contrast, the largest open-source model, Qwen3-VL-235B, struggles with an Execution MAE of 0.72 and a Consistency MAE of 0.91 (overall 0.81). This indicates that the image editing capabilities of current open-source models leave substantial room for improvement. However, after being trained on FIRM-Edit-370k, our FIRM-Edit-8B achieves the lowest Execution MAE of 0.53 and a Consistency MAE of 0.73. This yields an overall MAE of 0.62, successfully surpassing GPT-5 and all other open-source baselines. 

On FIRM-Bench-Gen, Gemini-3-Pro achieves the lowest MAE of 0.40, and GPT-5 achieves an MAE of 0.52, whereas the best open-source model, Qwen3-VL-32B, only reaches 0.54. Notably, despite having significantly fewer parameters, our FIRM-Gen-8B achieves a highly competitive MAE of 0.51, surpassing both GPT-5 and all evaluated open-source models. These results on FIRM-Bench directly demonstrate the effectiveness of our proposed framework.

  

\input{sections/tabs/tab_main_edit_rl}
\subsection{Image Editing Performance via RL process}

For the image editing task, we conduct experiments on Qwen-Image-Edit-2509~\cite{wu2025qwenimagetechnicalreport}, and report the post-RL evaluation results on GEditBench~\cite{liu2025step1x} and ImgEdit~\cite{ye2025imgedit} benchmarks in \cref{tab:editrl}.  As demonstrated, policy optimization guided by our FIRM-Edit-8B yields exceptional performance gains. Through merely a single RL stage, our FIRM-Qwen-Edit establishes a new SOTA score of 7.84 on GEditBench, while securing the second-highest score of 4.42 on ImgEdit. Remarkably, our model requires only $150 \times 16 = 2,400$ training samples to achieve performance comparable to UniWorld-Qwen-Image-Edit, which utilizes a much larger dataset of 27K samples. This stark contrast further underscores the  efficiency of our proposed method.

Notably, compared to utilizing the general-purpose Qwen3-VL-8B and Qwen3-VL-32B as reward models, FIRM-Edit-8B delivers a robust performance increase across both benchmarks (+0.30 on GEditBench and +0.07 on ImgEdit), which can be attributed to its highly accurate reward signals. In contrast, RL with Qwen3-VL-8B yields only minor improvements (+0.15 on GEditBench and +0.01 on ImgEdit). Furthermore, while Qwen3-VL-32B brings a slight gain of +0.11 on GEditBench, it surprisingly causes a performance drop of -0.07 on ImgEdit. This phenomenon strongly suggests that merely scaling up a general-purpose VLM does not guarantee better reward modeling. Instead, it highlights the indispensability of the precise and task-aligned reward signals provided by FIRM-Edit-8B, which are crucial for preventing negative optimization and guiding the generator toward superior editing fidelity. 
We present a comparison of the RL reward curves in \cref{fig:edit-reward-curves}. The rewards assigned by FIRM-Gen-8B are consistently lower than those from the Qwen3-VL series. This indicates that when evaluating image editing scenarios, the general-purpose Qwen3-VL models frequently overlook minor changes, resulting in artificially high scores.

\input{sections/tabs/tab_main_gen_rl}
\subsection{Image Generation Performance via RL process}
To evaluate generation capabilities, we employ SD3.5-Medium~\cite{rombach2022high} as the base model and report the post-RL evaluation results on the GenEval~\cite{ghosh2023geneval}, DPGBench~\cite{hu2024ella}, TIIF~\cite{wei2025tiif}(test-mini-short), and UniGenBench++~\cite{wang2025unigenbench++} benchmarks, as shown in \cref{tab:genrl}.
As demonstrated, FIRM-SD3.5, which was guided by FIRM-Gen-8B, yields profound performance gains across diverse and challenging settings. Our resulting FIRM-SD3.5 model achieves highly competitive scores of 0.77 on GenEval, 87.16 on DPGBench, 77.12 on TIIF, and 69.56/76.22 on UniGenBench-Short/Long. Remarkably, these results outperform heavily resourced models such as BAGEL and OmniGen2, which benefit from significantly larger parameter scales and training data.

Furthermore, in direct comparison with Qwen3-VL-8B and Qwen3-VL-32B acting as alternative reward models, FIRM-Gen-8B consistently drives superior generative outcomes, firmly validating the efficacy of our specialized reward formulation. On GenEval, which predominantly features short and unambiguous prompts, FIRM-SD3.5 provides a marginal advantage over the Qwen3-VL-8B baseline (+0.25 vs. +0.24). However, as the length and compositional complexity of the prompts escalate, the superiority of FIRM-Gen-8B becomes strikingly evident. Our model establishes substantial leads on DPGBench (+3.08 ours vs. +2.79), TIIF(+6.95 ours vs. +5.82), UniGenBench-Short/Long (+8.85/11.55 ours vs. +6.46/9.83). This robust scaling clearly demonstrates that our reward model excels particularly in guiding the generation of complex visual scenes.
A comparison of the RL reward curves for image generation is also depicted in \cref{fig:gen-reward-curves}. Interestingly, contrary to editing, the reward scores yielded by FIRM-Gen-8B are consistently higher than Qwen3-VL baselines. This phenomenon strongly suggests that in complex generative scenarios, the general-purpose models are prone to entangled evaluation criteria and severe hallucinations, which collectively lead to unjustifiably low scores.


\input{sections/tabs/tab_ablate_editmethod}
\subsection{Ablation on Reward Formulation}
To validate the effectiveness of our Consistency-Modulated Execution (CME) reward metric, we conduct an ablation study comparing several alternative reward formulations. The downstream evaluation results and their corresponding RL reward trajectories are detailed in \cref{tab:ablate_reward} and \cref{fig:rl-reward-curves-ablation}. It can be observed that, aside from our CME, all other baseline metrics suffer from various forms of performance degradation. Furthermore, the reward curves demonstrate that CME uniquely achieves a consistent and robust increase in reward signals throughout the optimization process.

For Edit-R1, utilizing Non-CoT logits provides noisy and confusing signals that fail to align with human preferences. Furthermore, relying on a simple weighted sum of scores leads to severe reward hacking. Specifically, the model can achieve a relatively high total reward by leaving the image completely unchanged; although the instruction-following score drops, it exploits the metric by securing a perfect appearance consistency score. This flawed reward structure incentivizes the editing model to become "lazy", opting to simply output the original input image, which ultimately results in poor editing performance. However, our CME metric effectively mitigates this issue. It consistently incentivizes the model to execute the given instruction while preserving task-irrelevant regions, thereby leading to significantly superior performance.

%% file: sections/tabs/tab_main_bench.tex
\begin{table}[tbp]
    \centering
    \begin{minipage}{0.48\textwidth}
        \centering
        \scriptsize
        \caption{Results on FIRM-Bench-Edit.}
        \label{tab:bench_generation}
        \begin{tabular}{lccc}
            \toprule
            \textbf{Model} & \textbf{Exec.} & \textbf{Cons.} & \textbf{Overall} \\
            \midrule
            GPT-4o & 0.73 & 1.16 & 0.93 \\
            GPT-4.1 & 0.74 & 1.04 & 0.88 \\
            GPT-5 & 0.62 & 0.73 & 0.67 \\
            Gemini-3-pro & \underline{0.54} & \textbf{0.57} & \textbf{0.55} \\
            \midrule
            InternVL3.5-8B & 0.90 & 1.13 & 1.00 \\
            InternVL3.5-38B & 0.85 & 1.22 & 1.02 \\
            InternVL3.5-241B-A28 & 0.69 & 1.05 & 0.86 \\
            Qwen3-VL-8B & 0.66 & 1.12 & 0.87 \\
            Qwen3-VL-32B & 0.69 & 1.14 & 0.90 \\
            Qwen3-VL-235B-A22 & 0.72 & 0.91 & 0.81 \\
            \midrule
            FIRM-Edit-8B (Ours) & \textbf{0.53} & \underline{0.73} & \underline{0.62} \\     
            \bottomrule
        \end{tabular}
    \end{minipage}%
    \hfill
    \begin{minipage}{0.50\textwidth}
        \centering
        \scriptsize
        \caption{Results on FIRM-Bench-Gen.}
        \label{tab:bench_edit}
        \begin{tabular}{lccc}
            \toprule
            \textbf{Model} & \textbf{Easy} & \textbf{Hard} & \textbf{Overall} \\
            \midrule
            GPT-4o & 0.63 & 0.66 & 0.65 \\
            GPT-4.1 & 0.58 & 0.69 & 0.64 \\
            GPT-5 & 0.50 & 0.53 & 0.52 \\
            Gemini-3-pro & \textbf{0.38} & \textbf{0.42} &  \textbf{0.40} \\
            \midrule
            InternVL3.5-8B & 0.61 & 0.55 & 0.58 \\
            InternVL3.5-38B & 0.79 & 0.58 & 0.68 \\
            InternVL3.5-241B-A28 & 0.60 & 0.61 & 0.60 \\
            Qwen3-VL-8B & 0.63 & 0.63 & 0.63 \\
            Qwen3-VL-32B & 0.51 & 0.56 & 0.54 \\
            Qwen3-VL-235B-A22 & 0.60 & \underline{0.52} & 0.56 \\
            \midrule
            FIRM-Gen-8B (Ours) & \underline{0.45} & 0.57 & \underline{0.51} \\
              
            \bottomrule
        \end{tabular}
    \end{minipage}
\end{table}

%% file: sections/tabs/tab_main_edit_rl.tex
\begin{table}[tbp]
  \centering
  \caption{Performance comparison on GEdit-Bench and ImgEdit. Guided by FIRM-Edit-8B during RL process, the resulting FIRM-Qwen-Edit substantially outperforms the base model as well as counterparts trained with Qwen3-VL models.}
  \label{tab:editrl}
  
  \resizebox{.9\textwidth}{!}{
  \tablestyle{14pt}{1.3}
    \begin{tabular}{lcccccccc}
      \toprule
      \multirow{2}{*}{\textbf{Model}}  & \multicolumn{3}{c}{\textbf{GEdit-Bench\cite{liu2025step1x}}} & \multicolumn{1}{c}{\textbf{ImgEdit\cite{ye2025imgedit}}}  \\
      \cmidrule(lr){2-4} \cmidrule(lr){5-5}  
        & G\_SC & G\_PQ & G\_Overall & Overall \\
      \midrule
      Instruct-Pix2Pix\cite{brooks2023instructpix2pix} & 3.58 & 5.49 & 3.68 & 1.88 \\
      AnyEdit\cite{yu2025anyedit} & 3.18& 5.82 & 3.21 & 2.45\\
      Step1X-Edit\cite{liu2025step1x} & 7.66& 7.35& 6.97& 3.06\\
      FLUX.1-Kontext[Dev]\cite{labs2025flux1kontextflowmatching} & 6.52& 7.38& 6.00& 3.71\\
      OmniGen2\cite{wu2025omnigen2} & 7.16 & 6.77 & 6.41 & 3.44\\
      FLUX.1-Kontext[Pro]\cite{labs2025flux1kontextflowmatching}  & 7.02 & 7.60 & 6.56 & 4.00 \\
      UniWorld-FLUX.1-Kontext\cite{li2025uniworld}  & 7.28& 7.49& 6.74& 4.02\\
      GPT-Image\cite{singh2025openai} & 7.85& 7.62& 7.53& 4.20\\
      UniWorld-Qwen-Image-Edit\cite{li2025uniworld} & 8.36& 7.87& \underline{7.76} & \textbf{4.48} \\
      \midrule
      Qwen-Image-Edit-2509\cite{wu2025qwenimagetechnicalreport} & 8.15 & 7.86& 7.54& 4.35\\
      - RL with Qwen3VL-8B & 8.04& 8.22& 7.69({\color{ForestGreen}+0.15})& 4.36({\color{ForestGreen}+0.01})\\
      - RL with Qwen3VL-32B & 7.94& 8.16& 7.65({\color{ForestGreen}+0.11})& 4.28({\color{red}-0.07})\\
      - \textbf{FIRM-Qwen-Edit (Ours)} &8.25& 8.20& \textbf{7.84}({\color{ForestGreen}\textbf{+0.30}}) & \underline{4.42}({\color{ForestGreen}\textbf{+0.07}})\\
      \bottomrule
    \end{tabular}
  }
\end{table}

%% file: sections/tabs/tab_main_gen_rl.tex
\begin{table}[tbp]
  \centering
  \caption{Performance comparison on GenEval, DPG, TIIF and UniGenBench++. By leveraging FIRM-Gen-8B for RL, our FIRM-SD3.5 achieves substantial performance gains, outperforming both baseline and counterparts trained with  Qwen3-VL models.}
  \label{tab:genrl}
  
  \resizebox{\textwidth}{!}{
  \tablestyle{14pt}{1.2}
    \begin{tabular}{lccccc}
      \toprule
      \multirow{2}{*} {\textbf{Model}} & 
      \multirow{2}{*} {\textbf{GenEval\cite{ghosh2023geneval}}} & 
      \multirow{2}{*} {\textbf{DPG-Bench\cite{hu2024ella}}} & 
      \multirow{2}{*} {\textbf{TIIF\cite{wei2025tiif}}} & 
      \multicolumn{2}{c}{\textbf{UniGenBench++\cite{wang2025unigenbench++}}} \\
      
      \cmidrule(lr){5-6}
      
      & & & & Short & Long \\
      \midrule
       SD-XL\cite{podell2023sdxl} & 0.55  & 74.65 & 54.96 & 40.22 & 41.48 \\
       Show-o\cite{xie2024showo} & 0.53 & - & 59.72 & - & - \\
       EMU3-Gen\cite{wang2024emu3} & 0.54 & 80.60 & - & 45.42 & 50.59 \\ 
       JanusFlow\cite{ma2025janusflow} & 0.63 & 79.68 & - & 47.10 & 54.80 \\
       FLUX.1-Dev\cite{flux2024}  & 0.66 & 83.84 & - & 60.97 & 69.42 \\
       DALLE-3\cite{hurst2024gpt}  & 0.67 & 83.50 & 74.96 & \underline{68.85} & 70.82 \\
       BLIP3o-4B\cite{chen2025blip3} & \underline{0.81} & 79.36 & - & 59.57 & 61.01 \\
       Janus-Pro-7B\cite{chen2025janus}& 0.80 & 84.19 & 66.50 & 61.36 & 71.11 \\
       Show-o2\cite{xie2025show} & 0.76 & 86.14 & - & 61.90 & 70.33 \\
       OmniGen2\cite{wu2025omnigen2} & 0.80 & 83.57 & - & 63.09 & 71.39 \\
       BAGEL\cite{deng2025emerging} & \textbf{0.82} & 85.07 & 71.50 & 59.91 & 71.26 \\
      \midrule
      SD3.5-Medium\cite{rombach2022high} & 0.52& 84.08 &70.17 & 60.71 & 64.67 \\
      - RL with Qwen3VL-8B  & 0.76({\color{ForestGreen}+0.24})& \underline{86.87}({\color{ForestGreen}+2.79}) &75.99({\color{ForestGreen}+5.82}) & 67.17({\color{ForestGreen}+6.46}) & \underline{74.50}({\color{ForestGreen}+9.83}) \\
      - RL with Qwen3VL-32B & 0.70({\color{ForestGreen}+0.18})& 85.94({\color{ForestGreen}+1.86})& \underline{76.43}({\color{ForestGreen}+6.26}) & 67.79({\color{ForestGreen}+7.08}) & 73.56({\color{ForestGreen}+8.89}) \\
      - \textbf{FIRM-SD3.5 (Ours)} & 0.77({\color{ForestGreen}\textbf{+0.25}})& \textbf{87.16}({\color{ForestGreen}\textbf{+3.08}}) &\textbf{77.12}({\color{ForestGreen}\textbf{+6.95}}) & \textbf{69.56}({\color{ForestGreen}\textbf{+8.85}}) & \textbf{76.22}({\color{ForestGreen}\textbf{+11.55}}) \\
      \bottomrule
    \end{tabular}
  }
\end{table}

%% file: sections/tabs/tab_ablate_editmethod.tex
\begin{table}[tbp]
  \centering
  \caption{Performance comparison of different reward formulations. Our proposed CME achieves the best overall performance by employing the "Base-and-Bonus" mechanism to effectively balance the competing rewards.}
  \label{tab:ablate_reward}
  
  \resizebox{\textwidth}{!}{
  \tablestyle{10pt}{1.3}
    \begin{tabular}{llccccccc}
      \toprule
      \multirow{2}{*}{\textbf{Method}}  &\multirow{2}{*}{\textbf{Formulation}} &\multirow{2}{*}{\textbf{Reward Model}} & \multicolumn{3}{c}{\textbf{GEdit-Bench}} & \multicolumn{1}{c}{\textbf{ImgEdit}}  \\
      \cmidrule(lr){4-6} \cmidrule(lr){7-7}  
        & & & G\_SC & G\_PQ & G\_Overall & Overall \\
      \midrule
      Baseline & - & - & 8.15 & 7.86& 7.54& 4.35\\
      Edit-R1 & Non-CoT Logits & Qwen2.5VL-32B & 4.08 & 5.05& 4.06 & 2.75\\
      Weighted Score & 0.5 * Exe. + 0.5 *Cons. & FIRM-Edit-8B &0.94& 8.71& 1.06& 2.17\\
      Weighted Score & 0.6 * Exe. + 0.4 *Cons. & FIRM-Edit-8B & 6.78& 8.25& 6.51& 3.73\\
      \textbf{CME (Ours)} & Exe. * (0.6 + 0.4 * Cons.) & FIRM-Edit-8B  &8.25& 8.20& \textbf{7.84} & \textbf{4.42}\\
      \bottomrule
    \end{tabular}
  }
\end{table}

\begin{figure}[t]
  \centering
  \begin{minipage}[t]{0.31\linewidth}
    \centering
    \includegraphics[width=\linewidth]{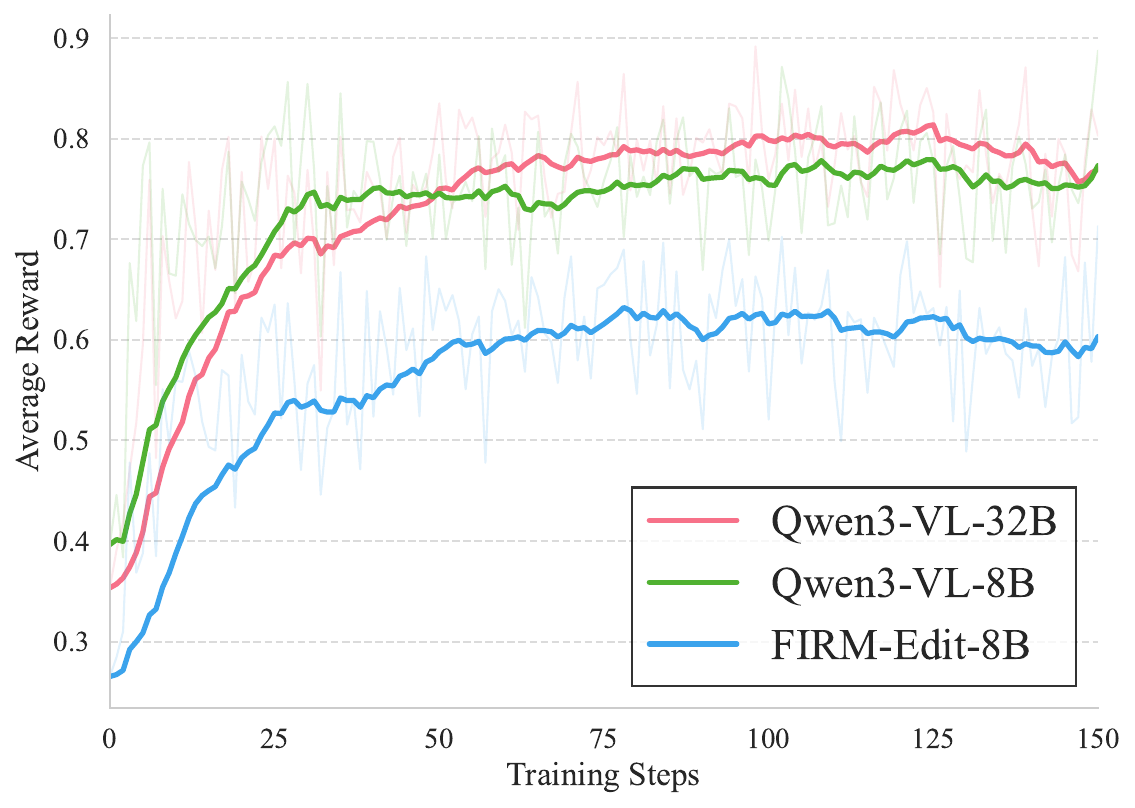}
    \caption{Editing RL reward curves of different reward models.}
    \label{fig:edit-reward-curves}
  \end{minipage}\hspace{0.02\linewidth}
  \begin{minipage}[t]{0.31\linewidth}
    \centering
    \includegraphics[width=\linewidth]{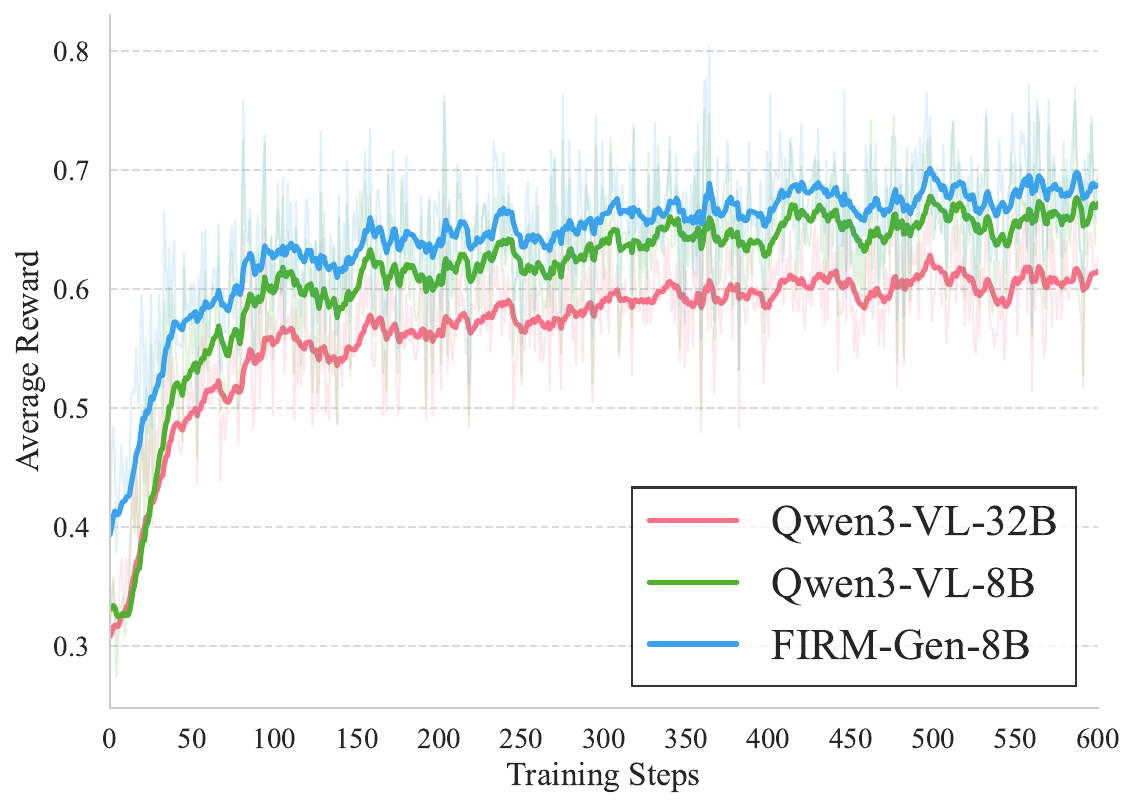}
    \caption{Generation RL reward curves of different reward models.}
    \label{fig:gen-reward-curves}
  \end{minipage}\hspace{0.02\linewidth}
  \begin{minipage}[t]{0.31\linewidth}
    \centering
    \includegraphics[width=\linewidth]{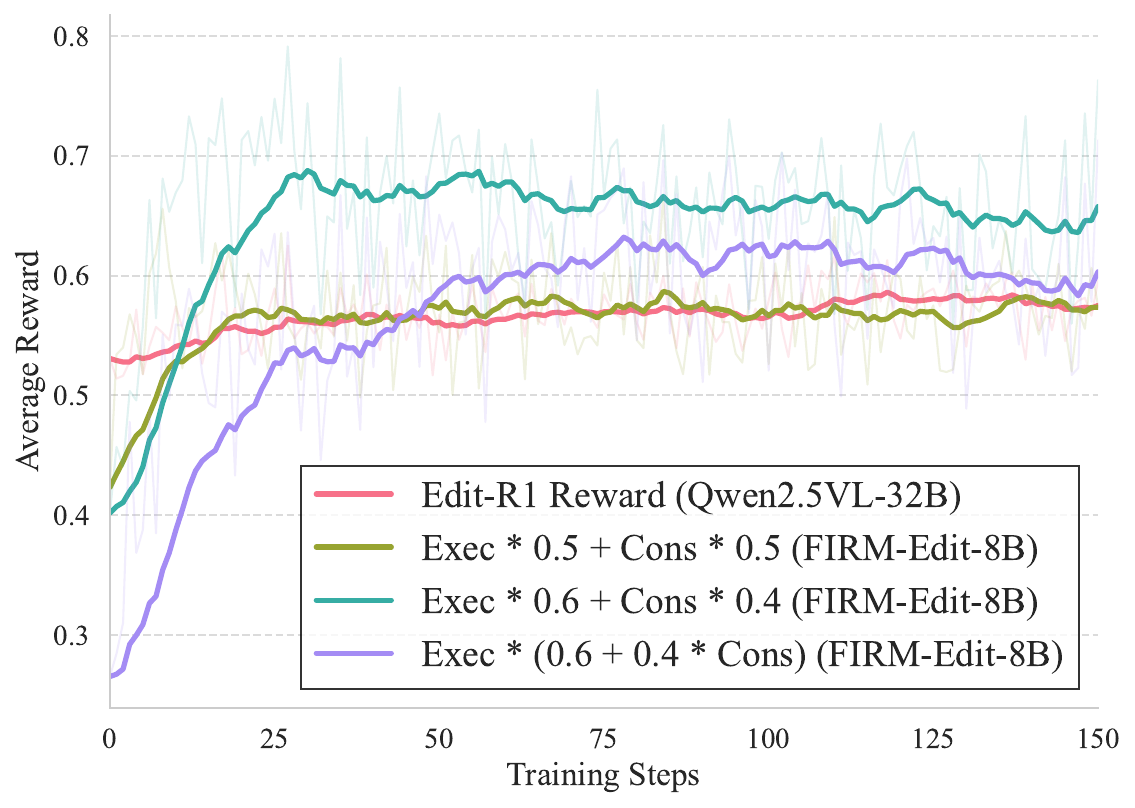}
    \caption{RL reward curves for the ablation of different reward calculation metrics.}
    \label{fig:rl-reward-curves-ablation}
  \end{minipage}
  
\end{figure}

%% file: sections/5_conclusion.tex
\section{Conclusion}
In this work, we present FIRM, a comprehensive framework designed for faithful image editing and generation. Our contributions encompass the entire reinforcement learning ecosystem, including tailored data curation pipelines, high-quality reward datasets, a rigorous evaluation benchmark, robust reward models, and novel reward formulations. 
By constructing this end-to-end pipeline, we successfully leverage our reward models to guide the RL process, achieving substantial performance improvements in downstream generative tasks. Ultimately, our study validates the indispensable role of accurate critics in the RL process, which we hope will inspire future research on reward-guided alignment for generative models.

%% file: sections/appendix.tex
\appendix

\section{Statistics Analysis of the FIRM Dataset and Benchmark}

\begin{table}[htbp]
\centering
\caption{Sample counts across score levels for the FIRM datasets and benchmarks.}
\label{tab:score_distribution}
\resizebox{\textwidth}{!}{
\renewcommand{\arraystretch}{1.2}   
\setlength{\tabcolsep}{12pt} 

\begin{tabular}{@{} l cccccc @{}}
\toprule
\multirow{2.5}{*}{\textbf{Score}} & \multicolumn{3}{c}{\textbf{Dataset}} & \multicolumn{3}{c}{\textbf{Benchmark}} \\
\cmidrule(lr){2-4} \cmidrule(lr){5-7}
& \textbf{Edit-Exec} & \textbf{Edit-Cons} & \textbf{Gen-Ins} & \textbf{Edit-Exec} & \textbf{Edit-Cons} & \textbf{Gen-Ins} \\
\midrule
5     & 49995 & 50823 & 116333  & 105 & 73 & 67 \\
4     & 16004 & 45720 & 41833  & 57 & 83 & 73 \\
3     & 52661 & 30999 & 56157 & 108 & 82 & 50 \\
2     & 21695 & 38486 & 59010  & 21 & 14 & 38 \\
1     & 49819  & 14053 & 19459  & 10  & 4 & 22 \\
\midrule
\textbf{Total} & \textbf{190174} & \textbf{180102} & \textbf{292792} & \textbf{301} & \textbf{256} &\textbf{250} \\
\bottomrule
\end{tabular}
}
\end{table}

We analyzed the sample distribution across different score ranges for both the FIRM dataset and the benchmark, as shown in \cref{tab:score_distribution}

\section{Analysis of Reward Function Hacking}
\begin{figure}[ht]
  \centering
  \includegraphics[width=0.48\textwidth]{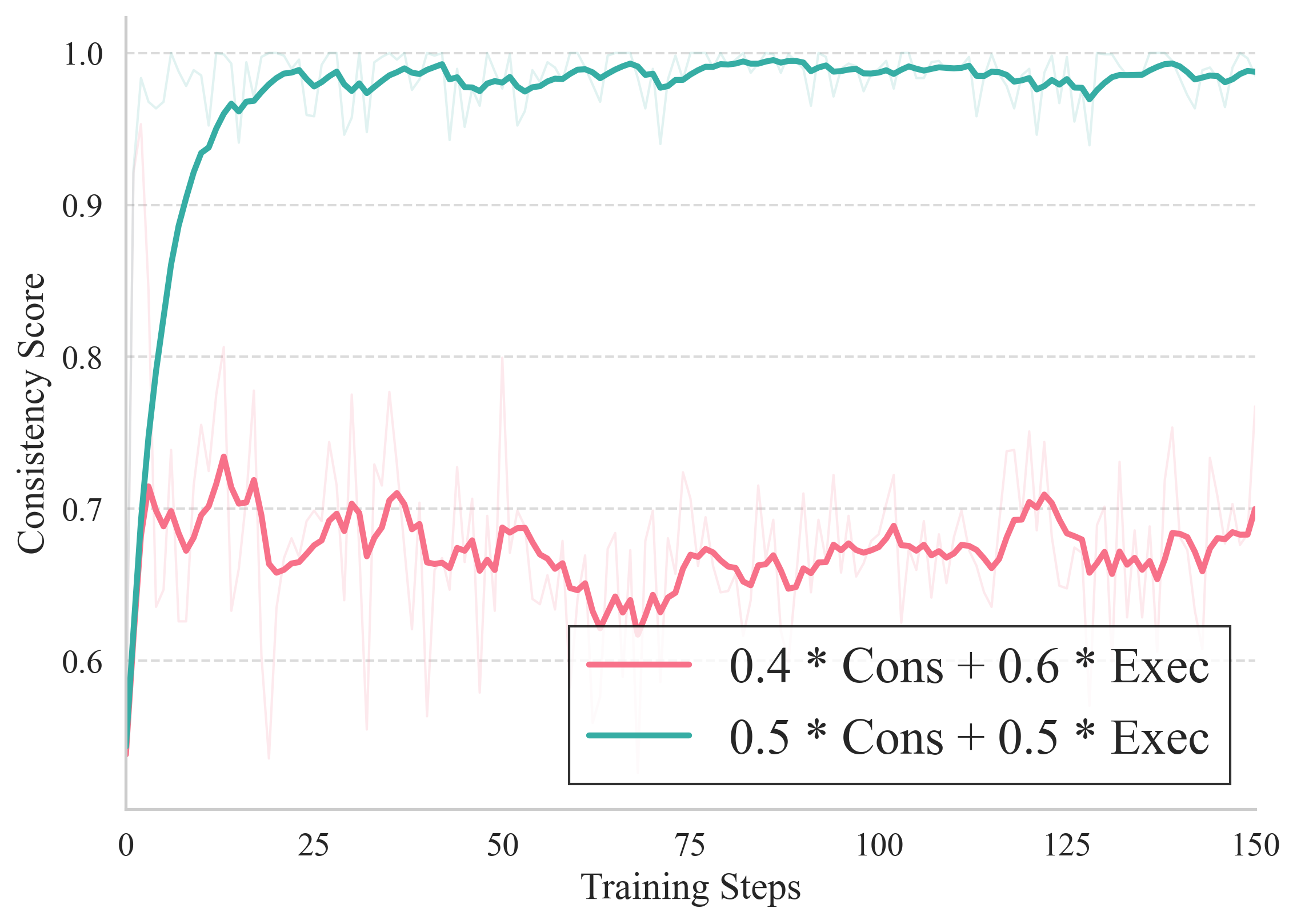}\hfill
  \includegraphics[width=0.48\textwidth]{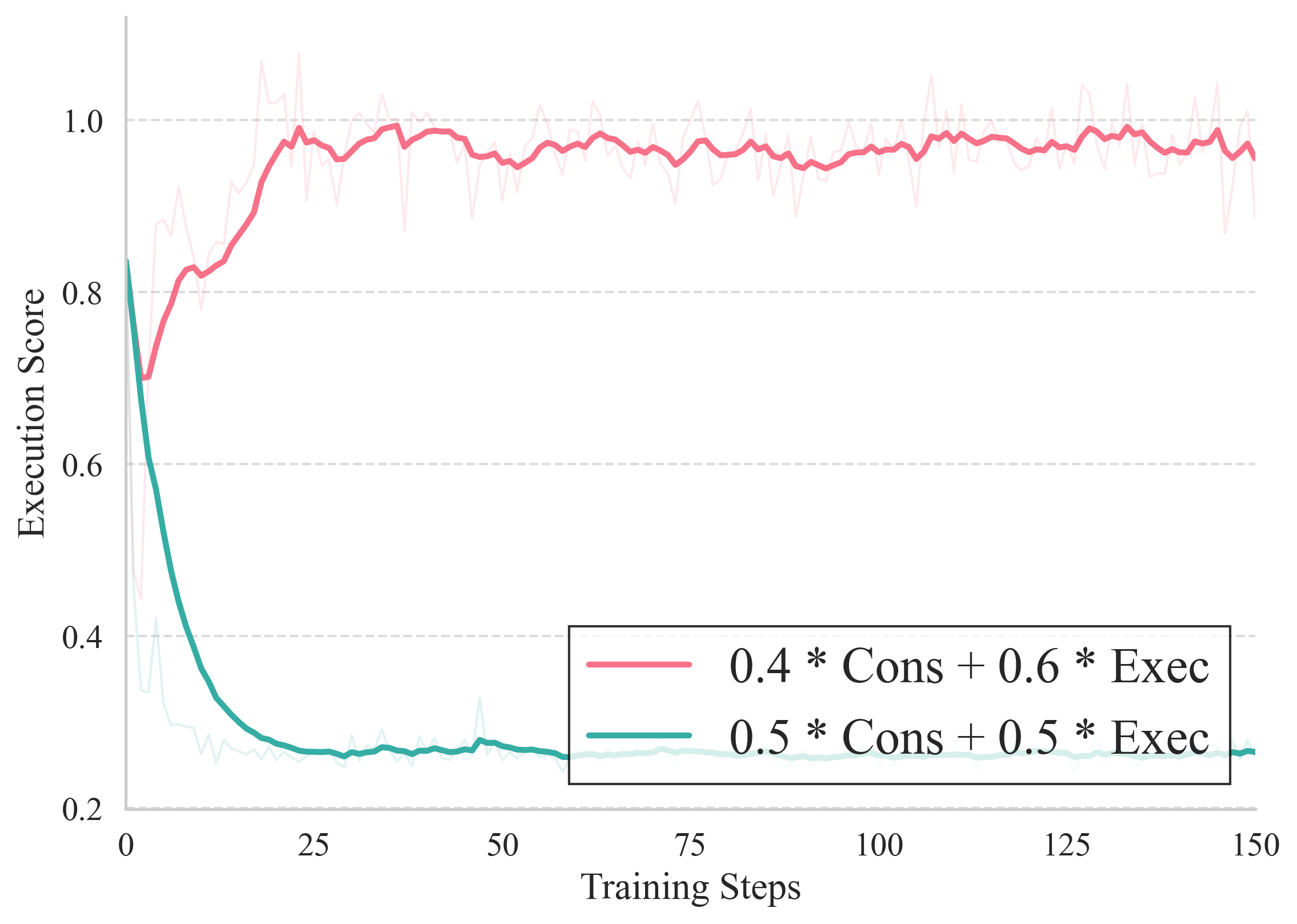}
  \caption{Reward Cureve of Consistency (left) and Execution (right) under raw reward designs.}
  \label{fig:cons-exec-curves}
\end{figure}

We initially define the reward as
\[
R = 0.5\,\text{Consistency} + 0.5\,\text{Execution}.
\]
However, this objective leads to severe reward hacking: the model over-optimizes \emph{Consistency} while neglecting \emph{Execution}. During RL training, it tends to reproduce outputs that are visually close to the source image, which yields high Consistency scores but poor task execution.

To mitigate this issue, we adjust the linear weighting to
\[
R = 0.4\,\text{Consistency} + 0.6\,\text{Execution}.
\]
This change partially improves execution performance, but the model then under-emphasizes consistency with the source image. As shown in Figure~\ref{fig:cons-exec-curves}, optimizing a fixed weighted sum still induces an undesirable trade-off between the two objectives.

We therefore propose \textbf{CME} with a multiplicative coupling:
\[
R = \text{Execution} \times \bigl(0.6 + 0.4\,\text{Consistency}\bigr).
\]
This formulation encourages strong execution while preserving alignment with the source image.

\section{Prompts for FIRM-Edit pipeline}
Here we present the prompts for each stage in FIRM-Edit data curation pipeline.

\begin{center}
\begin{AIbox}{Prompt for obvious editing differences}
\small\ttfamily
You are a highly skilled image comparator. You will receive two images. 

Task: Identify the obvious differences between the two images. Focus only on describing what subject has been changed and its own characteristics in the first and second images. Do NOT compare directly at this stage.
\end{AIbox} 
\label{fig:prompt_edit_diff_obvious}
\end{center}

\begin{center}
\begin{AIbox}{Prompt for detail editing differences}
\small\ttfamily
You are a highly skilled image comparator. You will receive two images. 

Task: Carefully compare ALL the differences between the two images in detail, including objects' Quantity, Shape \& Form, Texture \& Surface, Position \& Layout, etc.
\end{AIbox} 
\label{fig:prompt_edit_diff_detail}
\end{center}

\begin{center}
\begin{AIbox}{Prompt for Evaluation of Editing Execution}
\small\ttfamily
You are a highly skilled image evaluator. You will receive two images (an original image and a modified image), a description of how the edited image deviates from the original, and a specific edit instruction. The second image is known to have been altered based on this instruction, starting from the first image. Your task is to evaluate the execution successfulness of the edit instruction.

Task

Evaluate the execution successfulness of the edited image according to the following scale (1 to 5):

- 5 (Perfect Execution): The edited image perfectly implements all aspects of the instruction. All requested changes are present and correctly executed.

- 4 (Good Execution): The edited image successfully implements all key aspects of the instruction, with only a very subtle missing detail that doesn't significantly affect whether the instruction was followed.

- 3 (Partial Execution): The edited image implements the main intent of the instruction, but one significant element that was explicitly requested is missing or incorrectly implemented.

- 2 (Poor Execution): The edited image barely follows the instruction. Most requested changes are missing or incorrectly implemented, though there may be a vague attempt at following the instruction.

- 1 (No Execution): The edited image does not follow the instruction at all. No requested changes are visible, or the changes are completely contrary to what was requested.

CRITICAL - Evaluation Scope: 

- Only evaluate whether the REQUESTED changes are present and correctly implemented.

- Ignore any extra/unrequested modifications, rendering quality, realism, or unrelated consistency issues.

Output Format

You have to give your output in this way (Keep your reasoning concise and short.):

\{\{

"reasoning" : "<YOUR\_REASONING>",

"score" : [1/2/3/4/5]

\}\}

Input

Here is the text description of how the edited image deviates from the original image:

<START\_OF\_DIFFERENCE\_DESCRIPTION>

\{text\_context\}

<END\_OF\_DIFFERENCE\_DESCRIPTION>

Now evaluate how well the edited image follows the edit instruction:

<START\_OF\_EDIT\_INSTRUCTION>

\{edit\_prompt\}

<END\_OF\_EDIT\_INSTRUCTION>
\end{AIbox}
\label{fig:prompt_edit_execution}
\end{center}

\begin{center}
\begin{AIbox}{Prompt for Evaluation of Editing Consistency}
\small\ttfamily
You are a highly skilled image evaluator. You will receive two images (an original image and a modified image), a description of how the edited image deviates from the original, and a specific edit instruction. The second image is known to have been altered based on this instruction, starting from the first image. Your task is to evaluate how well the second image is consistent with the original image.

Definitions

- Significant Change: A noticeable alteration that substantially affects the visual perception or semantic content of the image.

- Minor Change: A subtle alteration that has limited impact on overall visual perception. 

Task

Evaluate the consistency between the images according to the following scale (1 to 5):

- 5: ONLY the changes explicitly required by the instruction are present. All other details are completely identical between the two images.

- 4: Besides changes explicitly required by the instruction, the second image contains 1 significant unintended change AND/OR 1-2 minor unintended changes.

- 3: Besides changes explicitly required by the instruction, the second image has 2-3 significant unintended changes AND/OR 3-4 minor unintended changes.

- 2: Besides changes explicitly required by the instruction, the second image has 4+ significant unintended changes AND/OR 5+ minor unintended changes.

- 1: The second image is almost entirely different from the original.

Requirements

CRITICAL- What Consistency Means:

- Consistency ONLY evaluates: "Did any changes occur that were NOT mentioned in the instruction?"

- It does NOT evaluate whether the instruction was successfully executed (that is evaluated separately).

Exceptions - Do NOT count as inconsistencies:

- Occlusion effects: Elements appearing/disappearing as a natural consequence of the instructed edit (e.g., background revealed when object is removed).

- Image quality variations: Small differences in sharpness, blur, noise, contrast, color temperature, lighting, reflection, shadow, saturation, etc. unless the instruction explicitly addresses these attributes.

- Entity Replacement EXPLICITLY instructed by instruction: When the instruction explicitly requires REPLACING entity A with B, ALL attributes of the new entity B are NOT consistency issues — only evaluate whether OTHER elements (background, other objects, scene composition) remain unchanged. NOTE: For ADD/REMOVE instructions, unintended entity removals/additions ARE inconsistencies. For Attribute Modification (e.g., change color, size, position), ONLY the specified attribute may change, any other changes in attributes of the same entity are inconsistencies.

- Environmental changes: Environmental changes that are a DIRECT PHYSICAL consequence of the instructed edit (e.g., lights turning on when changing daytime to night, wet ground when adding rain, shadows changing when lighting changes). Note: This does NOT include material substitutions/texture or object reposition/replacements that are merely aesthetically associated with the instruction.

Note: Apart from the exceptions listed above, other changes not explicitly instructed should be counted as inconsistencies."

Output Format

You have to give your output in this way (Keep your reasoning concise and short.):

\{\{

"reasoning" : "<YOUR\_REASONING>",

"score" : [1/2/3/4/5]

\}\}

Input

Here is the text description of how the edited image deviates from the original image:

<START\_OF\_DIFFERENCE\_DESCRIPTION>

\{text\_context\}

<END\_OF\_DIFFERENCE\_DESCRIPTION>

Now evaluate how well the edited image follows the edit instruction:

<START\_OF\_EDIT\_INSTRUCTION>

\{edit\_prompt\}

<END\_OF\_EDIT\_INSTRUCTION>

\end{AIbox}
\label{fig:prompt_edit_consistency}
\end{center}

\begin{center}
\begin{AIbox}{Prompt for Inferring Editing Execution Judgment}
\small\ttfamily
You are a highly skilled image evaluator. You will receive two images (an original image and a modified image) and a specific edit instruction. The second image is known to have been altered based on this instruction, starting from the first image. Your task is to evaluate the execution successfulness of the edit instruction.

Task

Evaluate the execution successfulness of the edited image according to the following scale (1 to 5):

- 5 (Perfect Execution): The edited image perfectly implements all aspects of the instruction. All requested changes are present and correctly executed.

- 4 (Good Execution): The edited image successfully implements all key aspects of the instruction, with only a very subtle missing detail that doesn't significantly affect whether the instruction was followed.

- 3 (Partial Execution): The edited image implements the main intent of the instruction, but some significant elements that was explicitly requested is missing or incorrectly implemented.

- 2 (Poor Execution): The edited image barely follows the instruction. Most requested changes are missing or incorrectly implemented, though there may be a vague attempt at following the instruction.

- 1 (No Execution): The edited image does not follow the instruction at all. No requested changes are visible, or the changes are completely contrary to what was requested.

CRITICAL - Evaluation Scope: 

- Only evaluate whether the REQUESTED changes are present and correctly implemented.

- Ignore any extra/unrequested modifications, rendering quality, realism, or unrelated consistency issues.

Output Format

You have to give your output in this way (Keep your reasoning concise and short.):

\{\{

"reasoning" : "<YOUR\_REASONING>",

"score" : [1/2/3/4/5]

\}\}

Input

Evaluate the execution successfulness of the edited image according to the edit instruction:

<START\_OF\_EDIT\_INSTRUCTION>

\{edit\_prompt\}

<END\_OF\_EDIT\_INSTRUCTION>
\end{AIbox}
\label{fig:prompt_edit_infer}
\end{center}

\begin{center}
\begin{AIbox}{Prompt for Inferring Editing Consistency Judgment}
\small\ttfamily
You are a highly skilled image evaluator. You will receive two images (an original image and a modified image) and a specific edit instruction. The second image is known to have been altered based on this instruction, starting from the first image. Your task is to evaluate how well the second image is consistent with the original image.

Definitions

Significant Change: A noticeable alteration that substantially affects the visual perception or semantic content of the image. 

Minor Change: A subtle alteration that has limited impact on overall visual perception. 

Task

Evaluate the consistency between the images according to the following scale (1 to 5):

- 5: ONLY the changes explicitly required by the instruction are present. All other details are completely identical between the two images.

- 4: Besides changes explicitly required by the instruction, the second image contains 1 significant unintended change AND/OR 1-2 minor unintended changes.

- 3: Besides changes explicitly required by the instruction, the second image has 2-3 significant unintended changes AND/OR 3-4 minor unintended changes.

- 2: Besides changes explicitly required by the instruction, the second image has 4+ significant unintended changes AND/OR 5+ minor unintended changes.

- 1: The second image is almost entirely different from the original.

Requirements

CRITICAL - What Consistency Means: 

- Consistency ONLY evaluates: "Did any changes occur that were NOT mentioned in the instruction?"

- It does NOT evaluate whether the instruction was successfully executed (that is evaluated separately).

Exceptions - Do NOT count as inconsistencies:

- Occlusion effects: Elements appearing/disappearing as a natural consequence of the instructed edit (e.g., background revealed when object is removed).

- Image quality variations: Small differences in sharpness, blur, noise, contrast, color temperature, lighting, reflection, shadow, saturation, etc. unless the instruction explicitly addresses these attributes.

- Entity Replacement EXPLICITLY instructed by instruction: When the instruction explicitly requires REPLACING entity A with B, ALL attributes of the new entity B are NOT consistency issues — only evaluate whether OTHER elements (background, other objects, scene composition) remain unchanged. NOTE: For ADD/REMOVE instructions, unintended entity removals/additions ARE inconsistencies. For Attribute Modification (e.g., change color, size, position), ONLY the specified attribute may change, any other changes in attributes of the same entity are inconsistencies.

- Environmental changes: Environmental changes that are a DIRECT PHYSICAL consequence of the instructed edit (e.g., lights turning on when changing daytime to night, wet ground when adding rain, shadows changing when lighting changes). Note: This does NOT include material substitutions/texture or object reposition/replacements that are merely aesthetically associated with the instruction.

Note: Apart from the exceptions listed above, other changes not explicitly instructed should be counted as inconsistencies."

Output Format

You have to give your output in this way (Keep your reasoning concise and short.):

\{\{

"reasoning" : "<YOUR\_REASONING>",

"score" : [1/2/3/4/5]

\}\}

Input

Evaluate how well the edited image is consistent with the original image given the edit instruction:

<START\_OF\_EDIT\_INSTRUCTION>

\{edit\_prompt\}

<END\_OF\_EDIT\_INSTRUCTION>
\end{AIbox}
\label{fig:prompt_edit_infer2}
\end{center}

\section{Prompts for FIRM-Gen pipeline}
Here we present the prompts for each stage in FIRM-Gen data curation pipeline.

\begin{center}
\begin{AIbox}{Prompt for Generating Analysis Plan}
\small\ttfamily
You are an expert visual QA analyst.

Your task is to analyze a text-to-image prompt and create a structured "Analysis Plan" text.

Tasks

1. Analyze the User Prompt deeply.

2. Break it down into verifiable visual criteria.

Output Requirement

output ONLY the analysis plan in the following text format (Markdown). Do NOT use JSON.

Analysis Plan Template

Analysis Plan:

To evaluate the image based on the given prompt, we need to break down the elements into specific questions and consider whether the image effectively meets each criterion:

1. [Main Subject/Element]:

 - [Question 1]?
 
 - [Question 2]?
 
2. [Attribute/Action]:

 - [Question 1]?
 
3. [Style/Composition]:

 - [Question 1]?
 
...

 - [Last Item]. [Negative Constraints](Optional, Include Only If Necessary):

 - Are any forbidden elements present (e.g., unwanted text, extra objects)?

For each of these points, you can provide a brief analysis indicating whether the image meets or exceeds expectations for each element.

User Prompt:

<START\_OF\_GENERATION\_INSTRUCTION>

\{generation\_prompt\}

<END\_OF\_GENERATION\_INSTRUCTION>

Output (Plan Text):
\end{AIbox}   
\label{fig:prompt_gen_plan}
\end{center}

\begin{center}
\begin{AIbox}{Prompt for Instruction Following Scoring}
\small\ttfamily
You are an expert Image Evaluator. 

Your task is to evaluate a generated image based on the Original Prompt and the Analysis Plan.

Tasks

1. Before writing, carefully inspect the image in full. Do not rush.

2. Then perform a step-by-step evaluation against the plan and provide ratings according to the rating scale below.

Rating Scale

- 5: All requirements, details, styles, and negative constraints are correct.

- 4: Main content is correct, but 1-2 non-critical details and requirements are slightly off.

- 3: Main subject(s) is present, but multiple requirements and details are missing.

- 2: The majority of main subject(s) are missing or incorrect, though a small portion of the content remains relevant.

- 1: Image is irrelevant to the original prompt.

Output Format

Produce the output in plain text, strictly following the structure below:

Begin with:

Let's evaluate the image against each element of the provided prompt:

1. [Criterion Name from Plan]:

- [Analysis...]

2. [Criterion Name from Plan]:

- [Analysis...]

...

(Analyze all points in the plan)

Final Analysis:

[A concise summary paragraph explaining the final decision and why the specific rating was chosen.]

Final Alignment Rating: [Rating]

\textbackslash\textbackslash boxed\{[Rating]\}

Constraints

1. The [Rating] inside \textbackslash\textbackslash boxed\{\} must be one of: 5, 4, 3, 2, 1.

2. Maintain objectivity. Treat the plan as a strict checklist and evaluate each criterion accordingly.

Original Prompt:

<START\_OF\_GENERATION\_INSTRUCTION>

\{generation\_prompt\}

<END\_OF\_GENERATION\_INSTRUCTION>

Analysis Plan:

<START\_OF\_ANALYSIS\_PLAN>

\{analysis\_plan\}

<END\_OF\_ANALYSIS\_PLAN>

Your Evaluation:
\end{AIbox}
\label{fig:prompt_gen_score}
\end{center}

\begin{center}
\begin{AIbox}{Prompt for Inferring Instruction Following Judgment}
\small\ttfamily
You are an expert Image Evaluator. 

Your task is to evaluate a generated image strictly based on the Original Prompt.

Tasks

1. Before writing, carefully inspect the image in full. Do not rush.

2. Identify all explicit and implicit requirements from the Original Prompt. This includes, but is not limited to, elements such as main subjects, attributes, actions, relationships, style, composition, and any negative constraints.

3. Perform a step-by-step evaluation by assessing whether the image satisfies each identified requirement.

4. Assign a final alignment rating according to the rating scale below.

Rating Scale

- 5: All requirements, details, styles, and negative constraints are correct.

- 4: Main content is correct, but 1-2 non-critical details and requirements are slightly off.

- 3: Main subject(s) is present, but multiple requirements and details are missing.

- 2: The majority of main subject(s) are missing or incorrect, though a small portion of the content remains relevant.

- 1: Image is irrelevant to the original prompt.

Output Format

Produce the output in plain text, strictly following the structure below:

Begin with:

Let's evaluate the image against the Original Prompt:

1. Identified Requirement 1:

- [Analysis...]

2. Identified Requirement 2:

- [Analysis...]

(Continue until all major requirements inferred from the prompt are evaluated)

Final Analysis:

[A concise summary paragraph explaining the final decision and why the specific rating was chosen.]

Final Alignment Rating: [Rating]

\textbackslash\textbackslash boxed\{[Rating]\}

Constraints

1. The [Rating] inside \textbackslash\textbackslash boxed\{\} must be one of: 5, 4, 3, 2, 1.

2. Maintain objectivity. Treat all identified requirements as a strict checklist and evaluate each one accordingly.

Original Prompt:

<START\_OF\_GENERATION\_INSTRUCTION>

\{generation\_prompt\}

<END\_OF\_GENERATION\_INSTRUCTION>
\end{AIbox}
\label{fig:prompt_gen_infer}
\end{center}